\documentclass[11pt]{article}
\usepackage[hyperref]{naaclhlt2019}
\usepackage{times}
\usepackage{url}
\usepackage{latexsym}
\usepackage{booktabs}
\usepackage{natbib}
\aclfinalcopy 

\title{Assessing BERT's Syntactic Abilities}

\author{Yoav Goldberg$^{1,2}$ \\
  $^1$ Computer Science Department, Bar Ilan University \\
  $^2$ Allen Institute for Artificial Intelligence \\
  {\tt yogo@cs.biu.ac.il , yoav@allenai.org}}

\date{}

\begin{document}
\maketitle
\begin{abstract}
I assess the extent to which the recently introduced BERT model captures English syntactic phenomena, using (1) naturally-occurring subject-verb agreement stimuli; (2) ``coloreless green ideas'' subject-verb agreement stimuli, in which content words in natural sentences are randomly replaced with words sharing the same part-of-speech and inflection; and (3) manually crafted stimuli for subject-verb agreement and reflexive anaphora phenomena. 
The BERT model performs remarkably well on all cases.
\end{abstract}

\section{Introduction}

The recently introduced BERT model \citep{bert} exhibits strong performance on several language understanding benchmarks. To what extent does it capture syntax-sensitive structures?

Recent work examines the extent to which RNN-based models capture syntax-sensitive phenomena that are traditionally taken as evidence for the existence in hierarchical structure. In particular, in \citep{linzen-goldberg} we assess the ability of LSTMs to learn subject-verb agreement patterns in English, and evaluate on naturally occurring wikipedia sentences. \cite{gulordava} also consider subject-verb agreement, but in a ``colorless green ideas'' setting in which content words in naturally occurring sentences are replaced with random words with the same part-of-speech and inflection, thus ensuring a focus on syntax rather than on selectional-preferences based cues.
\citet{marvin} consider a wider range of syntactic phenomena (subject-verb agreement, reflexive anaphora, negative polarity items) using manually constructed stimuli, allowing for greater coverage and control than in the naturally occurring setting.

The BERT model is based on the ``Transformer'' architecture \citep{transformer}, 
which---in contrast to RNNs---relies purely on attention mechanisms, and does not have an explicit notion of word order beyond marking each word with its absolute-position embedding. This reliance on attention may lead one\footnote{Indeed, they led me.} to expect decreased performance on syntax-sensitive tasks compared to RNN (LSTM) models that do model word order directly, and explicitly track states across the sentence. Indeed, \citet{tran} finds that transformer-based models perform worse than LSTM models on the \citet{linzen-goldberg} agreement prediction dataset.
In contrast, \cite{tang} find that self-attention performs on par with LSTM for syntax sensitive dependencies in the context of machine-translation, and performance on syntactic tasks is correlated with the number of attention heads in multi-head attention.

I adapt the evaluation protocol and stimuli of \citet{linzen-goldberg}, \citet{gulordava} and \citet{marvin} to the bidirectional setting required by BERT, and evaluate the pre-trained BERT models (both the \textsc{Large} and the \textsc{Base} models).
Surprisingly (at least to me), the out-of-the-box models (without any task-specific fine-tuning) perform very well on all the syntactic tasks.

\section{Methodology}

I use the stimuli provided by \citep{linzen-goldberg,gulordava,marvin}, but change the experimental protocol to adapt it to the bidirectional nature of the BERT model. This requires discarding some of the stimuli, as described below. Thus, the numbers are not strictly comparable to those reported in previous work.

\subsection{Previous setups}
All three previous work use uni-directional language-model-like models.

\citet{linzen-goldberg} start with existing sentences from wikipedia that contain a present-tense verb. They feed each sentence word by word into an LSTM, stop right before the focus verb, and ask the model to predict a binary plural/singular decision (supervised setup) or compare the probability assigned by a pre-trained language model (LM) to the plural vs singular forms of the verb (LM setup).\footnote{The results in \citet{linzen-goldberg} suggest that the supervised setup performs much better than the LM setup. However, several more recent work, including \citep{gulordava}, show that the LM setup does get results which are very close to that of the supervised one.}
The evaluation is then performed on sentences with ``agreement attractors'' in which at there is at least one noun between the verb and its subject, and all of the nouns between the verb and subject are of the opposite number from the subject.

\citet{gulordava} also start with existing sentences. However, in order to control for the possibillity of the model learning to rely on ``semantic'' selectional-preferences cues rather than syntactic ones, they replace each content word with random words from the same part-of-speech and inflection. This results in ``coloreless green ideas'' nonce sentences. The evaluation is then performed similarly to the LM setup of \citet{linzen-goldberg}: the sentence is fed into a pre-traiend LSTM LM up to the focus verb, and the model is considered correct if the probability assigned to the correct inflection of the original verb form given the prefix is larger than that assigned to the incorrect inflection.

\citet{marvin} focus on manually constructed and controlled stimuli, that also emphasizes linguistic structure over selectional preferences. 
They construct minimal pairs of grammatical and ungrammatical sentences, feed each one \emph{in its entirety} into a pre-trained LSTM-LM, and compare the perplexity assigned by the model to the grammatical and ungrammatical sentences. The model is ``correct'' if it assigns the grammatical sentence a higher probability than to the ungrammatical one. Since the minimal pairs for most phenomena differ only in a single word (the focus verb), this scoring is very similar to the one used in the two previous works. However, it does consider the continuation of the sentence after the focus verb, and also allows for assessing phenomena that require change into two or more words (like negative polarity items).

\subsection{Adaptation to the BERT model}

In contrast to these works, the BERT model is bi-directional: it is trained to predict the identity of masked words based on both the prefix and suffix surrounding these words. I adapt the uni-directional setup by feeding into BERT the complete sentence, while \emph{masking out the single focus verb}. I then ask BERT for its word predictions for the masked position, and compare the score\footnote{I use the pre-softmax logit scores to save a bit of computation, but this is equivalent to using probabilities.} assigned to the original correct verb to the score assigned to the incorrect one. 

For example, for the sentence:\\[0.5em]
\emph{a 2002 systemic review of herbal products found that several herbs , including peppermint and caraway , \underline{have} anti-dyspeptic effects for non-ulcer dyspepsia with `` encouraging safety profiles '' .} (from \citep{linzen-goldberg})\\[0.5em] 
I feed into BERT:\\[0.5em]
\texttt{[CLS] a 2002 systemic review of herbal products found that several herbs , including peppermint and caraway , [MASK] anti-dyspeptic effects for non-ulcer dyspepsia with `` encouraging safety profiles '' .} and look for the score assigned to the words \texttt{have} and \texttt{has} at the masked position.

Similarly, for the pair\\[0.5em]
\emph{the game that the guard hates is bad .}\\
\emph{the game that the guard hates are bad .}\\ (from \cite{marvin}), I feed into BERT:\\[0.5em] \texttt{[CLS] the game that the guard hates [MASK] bad .}\\[0.5em] and compare the scores predicted for \texttt{is} and \texttt{are}.

This differs from \citet{linzen-goldberg} and \citet{gulordava} by considering the entire sentence (excluding the verb) and not just its prefix leading to the verb, and differs from \citet{marvin} by conditioning the focus verb on bidirectional context. 

I use the PyTorch implementation of BERT, with the pre-trained models supplied by Google.\footnote{\url{https://github.com/huggingface/pytorch-pretrained-BERT}}
I experiment with the \texttt{bert-large-uncased} and \texttt{bert-base-uncased} models.

\paragraph{Discarded Material} 
The bi-directional setup precludes using using the NPI stimuli of \citet{marvin}, in which the minimal pair differs in two words position, which I discard from the evaluation. I also discard the agreement cases involving the verbs \textit{is} or \textit{are} in \citet{linzen-goldberg} and in \citet{gulordava}, because some of them are copular construction, in which strong agreement hints can be found also on the object following the verb.\footnote{Results are generally a bit higher when not discarding the is/are cases.} This is not an issue in the manually constructed \citep{marvin} stimuli due to the patterns they chose.

Finally, I discard stimuli in which the focus verb or its plural/singular inflection does not appear as a single word in the BERT word-piece-based vocabulary (and hence cannot be predicted by the model).
This include discarding \citet{marvin} stimuli involving the words \emph{swims} or \emph{admires}, resulting in 23,368 discarded pairs (out of 152,300).
I similarly discard 680 sentences from \citep{linzen-goldberg} where the focus verb or its inflection were one of 108 out-of-vocabulary tokens,\footnote{blames, dislike, inhabit, exclude, revolves, governs, delete, composes, overlap, edits, embrace, compose, undertakes, disagrees, redirect, persist, recognise, rotates, accompanies, attach, undertake, earn, communicates, imagine, contradicts, specialize, accuses, obtain, caters, welcomes, interprets, await, communicate, templates, qualify, reverts, achieve, achieves, govern, restricts, violate, behave, emit, contend, adopt, overlaps, reproduces, rotate, defends, submit, revolve, lend, pertain, disagree, concentrate, detects, endorses, detect, predate, persists, consume, locates, earns, predict, interact, merge, consumes, behaves, locate, predates, enhances, predicts, integrates, inhabits, satisfy, contradict, swear, activate, restrict, satisfies, redirects, excludes, violates, interacts, admires, speculate, blame, drag, qualifies, activates, criticize, assures, welcome, depart, characterizes, defend, obtains, lends, strives, accuse, recognises, characterize, contends, perceive, complain, awaits} and 28 sentence-pairs (8 tokens\footnote{toss, spills, tosses, affirms, spill, melt, approves, affirm}) from \citep{gulordava}.

\paragraph{Limitations}
The BERT results are not directly comparable to the numbers reported in previous work. Beyond the differences due to bidirectionality and the discarded stimuli, the BERT models are also trained on a different and larger corpus (covering both wikipedia and books).

\paragraph{Reproducability}
Code is available at \url{https://github.com/yoavg/bert-syntax}.

\section{Results}

\begin{table}[t!]
\centering
\begin{tabular}{cccccc}
\toprule
Attractors & BERT Base & BERT Large & \# sents \\
\hline
1 & 0.97 & 0.97 & 24031 \\
2 & 0.97 & 0.97 & 4414 \\
3 & 0.96 & 0.96 & 946 \\
4 & 0.97 & 0.96 & 254 \\
\bottomrule
\end{tabular}
\caption{Results on the \citet{linzen-goldberg} stimuli. While not directly comparable, the numbers reported by \citet{linzen-goldberg} and \citet{gulordava} for the 2, 3 and 4 attractor cases are substantially lower.}
\label{tbl:nat}
\end{table}

\begin{table}[t!]
\centering
\begin{tabular}{cccccc}
\toprule
Attractors & BERT Base & BERT Large & \# pairs \\
\hline
All & 0.83 & 0.80 & 383 \\
\hline
0 & 0.84 & 0.80 & 311 \\
1 & 0.81 & 0.75 & 63 \\
2 & 0.89 & 0.89 & 9 \\
\bottomrule
\end{tabular}
\caption{Results on the \textsc{En Nonce} \citep{gulordava} stimuli. While not strictly comparable, the numbers reported by \citet{gulordava} for the LSTM in this condition (on All) is $74.1\pm1.6$.}
\label{tbl:cgi}
\end{table}

\begin{table*}[t!]
\centering
\begin{tabular}{lccccc}
\toprule
& BERT & BERT & LSTM & Humans &  \# Pairs \\
& Base & Large & (M\&L) & (M\&L) & (\# M\&L Pairs) \\
\hline
\textsc{Subject-verb agreement:} \\
Simple & 1.00 & 1.00 & 0.94 & 0.96 & 120 (140) \\
In a sentential complement & 0.83 & 0.86 & 0.99 & 0.93 & 1440 (1680) \\
Short VP coordination & 0.89 & 0.86 & 0.90 & 0.82 & 720 (840) \\
Long VP coordination & 0.98 & 0.97 & 0.61 & 0.82 & 400 (400) \\
Across a prepositional phrase & 0.85 & 0.85 & 0.57 & 0.85 & 19440 (22400) \\
Across a subject relative clause & 0.84 & 0.85 & 0.56 & 0.88 & 9600 (11200) \\
Across an object relative clause & 0.89 & 0.85 & 0.50 & 0.85 & 19680 (22400) \\
Across an object relative (no \textit{that}) & 0.86 & 0.81 & 0.52 & 0.82 & 19680 (22400)\\
In an object relative clause & 0.95 & 0.99 & 0.84 & 0.78 & 15960 (22400) \\
In an object relative (no \textit{that}) & 0.79 & 0.82 & 0.71 & 0.79 & 15960 (22400) \\
\hline
\textsc{Reflexive anaphora:} \\
Simple & 0.94 & 0.92 & 0.83 & 0.96 & 280 (280) \\
In a sentential complement & 0.89 & 0.86 & 0.86 & 0.91 & 3360 (3360) \\
Across a relative clause & 0.80 & 0.76 & 0.55 & 0.87 & 22400 (22400) \\
\bottomrule
%simple_npi_anim & 1.00 & 1.00 & 246 \\
\end{tabular}
\caption{Results on the \citet{marvin} stimuli. M\&L results numbers are taken from \citet{marvin}. The BERT and M\&L numbers are \emph{not} directly comparable, as the experimental setup differs in many ways.}
\label{tbl:synt}
\end{table*}

Tables \ref{tbl:nat}, \ref{tbl:cgi} and \ref{tbl:synt} show the results.
All cases exhibit high scores---in the vast majority of the cases substantially higher than reported in previous work.\footnote{The only exception are the ``In a sentential complement'' and ''Short VP coordination'' conditions in Table \ref{tbl:synt}. However, the much better scores of the BERT model on the other conditions in that table suggest that the high LSTM numbers on these conditions are due to overfitting to a particular construction, rather tan good syntactic generalization.} As discussed above, the results are \emph{not directly comparable to previous work}: the BERT models are trained on different (and larger) data, are allowed to access the suffix of the sentence in addition to its prefix, and are evaluated on somewhat different data due to discarding OOV items. Still, taken together, the high performance numbers indicate that the purely attention-based BERT models are likely capable of capturing the same kind of syntactic regularities that LSTM-based models are capable of capturing, at least as well as the LSTM models and probably better. 

Another noticeable and interesting trend is that \textbf{larger is not necessarily better}: the BERT-Base model outperforms the BERT-Large model on many of the syntactic conditions.

\section{Discussion}
The BERT models perform remarkably well on all the syntactic test cases. I expected the attention-based mechanism to fail on these (compared to the LSTM-based models), and am surprised by these results. The \citet{gulordava} and \citet{marvin} conditions rule out the possibility of overly relying on selectional preference cues or memorizing the wikipedia training data, and suggest real syntactic generalization is taking place. Exploring the \emph{extent} to which deep purely-attention-based architectures such as BERT are capable of capturing hierarchy-sensitive and syntactic dependencies---as well as the \emph{mechanisms} by which this is achieved---is a fascinating area for future research.

\bibliographystyle{acl_natbib}
\bibliography{refs}

\end{document}